\documentclass[10pt, a4paper]{article}
\usepackage{lrec-coling2024} 

\usepackage{natbib}
\usepackage{multibib}
\makeatletter
\def\@mb@citenamelist{cite,citep,citet,citealp,citealt,citepalias,citetalias}
\makeatother
\newcites{languageresource}{~}

\usepackage{graphicx}
\usepackage{tabularx}
\usepackage{soul}

\usepackage{multirow}
\usepackage{multicol}
\usepackage{booktabs}
\usepackage{svg}
\usepackage{amssymb} 

\usepackage{xcolor}
\usepackage{hyperref}
 \definecolor{darkblue}{rgb}{0, 0, 0.5}
  \hypersetup{colorlinks=true, citecolor=darkblue, linkcolor=darkblue, urlcolor=darkblue}

\usepackage{xstring}

\usepackage{color}

\title{Automatic Speech Recognition System-Independent \\Word Error Rate Estimation}

\name{Chanho Park, Mingjie Chen, Thomas Hain} 

\address{Speech and Hearing Research Group, University of Sheffield, UK \\
         \{cpark12, mingjie.chen, t.hain\}@sheffield.ac.uk\\}

\abstract{
Word error rate (WER) is a metric used to evaluate the quality of transcriptions produced by Automatic Speech Recognition (ASR) systems. In many applications, it is of interest to estimate WER given a pair of a speech utterance and a transcript. Previous work on WER estimation focused on building models that are trained with a specific ASR system in mind (referred to as ASR system-dependent). These are also domain-dependent and inflexible in real-world applications. In this paper, a hypothesis generation method for ASR System-Independent WER estimation (SIWE) is proposed. In contrast to prior work, the WER estimators are trained using data that simulates ASR system output. Hypotheses are generated using phonetically similar or linguistically more likely alternative words. In WER estimation experiments, the proposed method reaches a similar performance to ASR system-dependent WER estimators on in-domain data and achieves state-of-the-art performance on out-of-domain data. On the out-of-domain data, the SIWE model outperformed the baseline estimators in root mean square error and Pearson correlation coefficient by relative 17.58\% and 18.21\%, respectively, on Switchboard and CALLHOME. The performance was further improved when the WER of the training set was close to the WER of the evaluation dataset. 
 \\ \newline  }

\begin{document}

\maketitleabstract

\section{Introduction}
Automatic quality estimation (QE) of machine learning models has gained increasing popularity. Many types of quality estimation have been studied \cite{negri-etal-2014-quality, 10095888, kurz2022uncertainty, mittag20b_interspeech, zerva-etal-2022-findings}, where the aim is to estimate the quality of outputs of models. It was found to be useful for measuring model performance for practical tasks, such as image classification \cite{kurz2022uncertainty}, machine translation \cite{specia2018quality}, automatic speech recognition (ASR) \cite{10095888}, speech synthesis \cite{huang2022ldnet} and speech enhancement \cite{reddy2022dnsmos}, especially when there is no oracle references or ground-truth labels.

This paper mainly focuses on quality estimation of ASR system output. ASR research \cite{povey16_interspeech, graves2012sequence, NEURIPS2020_92d1e1eb, gulati20_interspeech, radford2022robust} has achieved impressive successes and made huge progress on transcription accuracy and computational efficiency. Even so, errors still occur and it remains essential to evaluate the quality of ASR transcripts independently at low cost, as oracle references are not always available, especially in real-world applications or production scenarios.

Several methods to estimate the quality of ASR transcripts have been proposed. As summarised in \citet{negri-etal-2014-quality}, these methods can be roughly categorised into two classes: method using glass-box and black-box features. For the glass-box methods, to estimate the quality of ASR outputs, intermediate features of ASR systems, such as confidence scores \cite{jiang2005confidence, kalgaonkar2015estimating} are typically used. Another type is a black-box method that directly estimates the word error rate (WER) of transcripts and audio \cite{negri-etal-2014-quality, ali-renals-2018-word,ali20_interspeech, 10095888, park2023fast}. The former methods using a confidence score have been used widely for different applications such as semi-supervised learning \cite{drugman2019active} and speaker-adaptation \cite{deng2023confidence}. However, a common issue of ASR systems derived from confidence scores is that they can be overconfident  \cite{li2021confidence}, which leads to low performance of quality estimation. Compared to confidence score-based methods, WER estimators using black-box features are not dependent on the intermediate features from specific ASR systems. Hence, the WER estimators can have computational efficiency advantages at inference because they do not require running expensive ASR decoding. However, a common issue of most previous WER estimators \cite{negri-etal-2014-quality, 10095888, park2023fast} is that their training is dependent on factors such as the nature of hypotheses or training datasets. These dependencies may be causes of performance degradation or result in a narrow range of use of WER estimators.

Typically, the training datasets of WER estimators are generated based on speech datasets with ground-truth references. These datasets are composed of pairs of a speech utterance and a hypothesis, as well as the target WER. The latter is derived by comparing a hypothesis with the reference transcript and computing the so-called edit distance. The hypotheses are generated using a specific ASR system. Hence, the WER estimators \cite{negri-etal-2014-quality, 10095888, park2023fast} are dependent on the ASR systems and the speech datasets used to generate the training datasets. In this paper, for simplicity, these WER estimators are named system-dependent WER estimators. When estimating the quality of output from another ASR system, system-dependent WER estimators do not perform well and need to be re-trained, which reduces the usefulness of ASR system-dependent WER estimators. In addition, they are also likely to suffer performance degradation on out-of-domain test data, where errors occur that previously have not been observed.

The work in this paper tries to address the issues above by proposing a System-Independent WER Estimation (SIWE) method. Instead of generating training datasets using ASR systems, a range of data augmentation methods are proposed that allow to generate plausible hypotheses. The data augmentation methods deduce hypotheses from ground-truth references by inserting errors. Three types of strategies are used for error insertion: producing insertion, deletion and substitution errors. The training datasets generated from the proposed methods help SIWE to achieve state-of-the-art WER estimation performance. On in-domain test data, SIWE reaches the same level as performance of the system-dependent WER estimators. Furthermore, SIWE outperformed the system-dependent WER estimators on out-of-domain test data.

The main contributions of this paper can be summarised as follows:
\begin{itemize}
	\item This paper proposes a System-Independent WER estimator, which enjoys a broader range of use than system-dependent WER estimators.
	\item This paper proposes a new data augmentation method for generating training datasets for WER estimators.
	\item The proposed SIWE model reaches the same level of performance as system-dependent estimators on in-domain test data and it further outperformed the system-dependent WER estimators on out-of-domain test data.
\end{itemize}

\begin{figure*}[htbp]
    \centering
    \includegraphics[width=\textwidth]{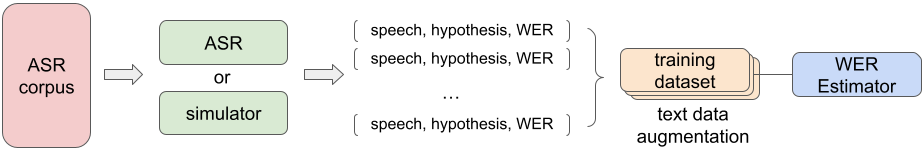}
    \caption{Illustration of WER estimation}
    \label{figure:Illustration of WER estimation}
\end{figure*}

\section{Related Works}
\subsection{Word Error Rate Estimation}
\citet{negri-etal-2014-quality} proposed an estimation method for ASR QE without manual reference transcripts. They trained a WER estimation model using both glass-box and black-box features. \citet{ali-renals-2018-word} also proposed a feed-forward neural network called e-WER for predicting the number of errors and word count per spoken utterance. In \citet{ali20_interspeech}, the features extracted from a phone recogniser were used to improve the performance. Moreover, e-WER3 \cite{10095888} extended the previous model to predict WER on multiple language data including English. In the study, only black-box features were employed. In addition to these methods, the Fast WER estimator (Fe-WER), has been proposed by \citet{park2023fast}. This model reduced the computational cost without performance degradation by adopting self-supervised learning representations for speech and text as black-box features. For evaluation, mean absolute error (MAE), root mean square error (RMSE) and Pearson correlation coefficient (PCC) were measured.

\subsection{Automatic Speech Recognition Systems}
\label{subsection:Automatic Speech Recognition Systems}
ASR has experienced impressive success and made huge progress in recent years~\cite{SIP-2021-0050}. They have significant differences in many ways, including model architectures, input features and training objectives. One type of ASR systems is based on so-called hybrid modelling~\cite{6638967, povey2014parallel}. The system consists of multiple independent modules and is trained through a training objective, such as lattice-free maximum mutual information (LF-MMI)~\cite{povey16_interspeech}. End-to-end (E2E) ASR systems have gained much attention. For E2E to be robust to the long context, recurrent neural network (RNN) transducer~\cite{graves2012sequence} has been adopted by \citet{8682336}. Its encoder is jointly trained with prediction networks that depend on previous labels. Moreover, Transformer has been integrated with E2E models: wav2vec 2.0~\cite{NEURIPS2020_92d1e1eb}; Conformer~\cite{gulati20_interspeech}; Whisper~\cite{radford2022robust}. First, wav2vec 2.0 is pre-trained to learn contextualised representation. Then, representation is linearly projected into output tokens. The model is optimised by minimising the connectionist temporal classiﬁcation loss. Second, Conformer is a model based on a transformer and convolutional neural networks for feature extraction. Last, Whisper is an ASR system trained on large amounts of transcripts collected from the internet. It is trained on multiple tasks, such as speech recognition and language identification.

\section{ASR System-Independent WER Estimator}
\subsection{WER Estimation}
Figure~\ref{figure:Illustration of WER estimation} depicts ASR system-independent WER estimation. The first step is generating training data for WER estimation. Spoken utterances can be transcribed by an ASR system for baseline models or a simulator for proposed models. A training instance for WER estimation consists of a speech utterance, a hypothesis and a WER between the hypothesis and the reference transcript. In previous studies~\cite{negri-etal-2014-quality, 10095888, park2023fast}, the hypotheses are produced from an ASR system, which, by implication, makes the WER estimation models dependent on the ASR system. This work proposes to derive hypotheses from reference transcripts via data generation. At training, the WER estimator is usually optimised as a regression model through the MSE loss function. 

\subsection{Fe-WER}
\subsubsection{Model Architecture}
\begin{figure}[htbp]
    \centering
    \includegraphics[width=6cm]{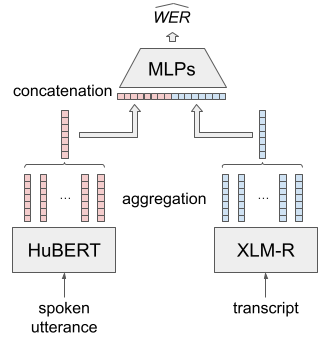}
    \caption{Overview of Fe-WER architecture}
    \label{figure:Overview of Fe-WER architecture}
\end{figure}

The WER estimator is built upon the method suggested by \citet{park2023fast}. The model is based on a two-tower architecture~\cite{10.1145/2505515.2505665}, where one tower is a speech representation model and the other is a text representation model as described in Figure~\ref{figure:Overview of Fe-WER architecture}. 

The tower models extract frame-level features from a spoken utterance and token-level features from a transcript. The features are averaged over frames or tokens. The aggregated features for speech and text are concatenated as an input to the multi-layer perceptrons (MLPs). Then, the MLPs output a WER estimate. 

\subsubsection{Training Objective}
The model is trained to minimise mean squared error (MSE). 
\begin{equation}
\label{equation:MSE}
    MSE = \frac{1}{N}\sum_{i=1}^{N} (WER - \widehat{WER})^2
\end{equation}
where $N$ is the number of instances, $WER$ is an actual WER and $\widehat{WER}$ is an estimate. 

\subsubsection{Evaluation Metrics}
RMSE is used as an evaluation metric for WER estimation as well as PCC. These metrics have been adopted for WER estimation in the recent studies~\cite{10095888, park2023fast}. RMSE measures the average difference between estimates and actual WERs, while PCC measures the relationship between them. PCC is from $-1$ to $1$ and if it is $1$, then the estimate tends to increase when the target increases, while it is $0$ if there is no relationship between them. 

\subsection{Hypothesis Generation}
\subsubsection{Hypothesis Generation Strategy}
There are three main strategies for hypothesis generation: random selection, phonetic similarity, linguistic probability. These approaches generate the errors of an ASR, an acoustic and a language model, respectively. 

\textbf{Random Selection Strategy} Positions for insertions, deletions and substitutions are selected randomly, aiming for a specific target WER. 

\textbf{Phonetic Similarity Strategy} An ASR model produces ASR errors between phonetically similar words, such as \textit{grief} and \textit{brief}. Thus, the phonetic similarity between the reference word and the other words can be considered when the substitution is generated. First of all, a word in a reference transcript is converted into a phoneme sequence, e.g. speech to \textit{S P IY CH}. Then, the edit distance between two words is calculated. After calculating all the distances to the other words in a vocabulary list, the top $n$ words similar to the reference word are listed. When a word is substituted with another word, the word for substitution is selected from the phonetic similar word list. 

\textbf{Linguistic Probability Strategy} In the case of insertion, it can be caused by grammatical corrections. Here is an example of the insertion related to the context rather than the acoustic similarity: [I do not] know if he is alive. The words in [] are inserted by an ASR model with a language model, which is used to get the probability of the word in context.

The relation between Hypothesis generation strategies and edit types is described in Table~\ref{table:Hypothesis generation strategies and edit types}.
\begin{table}[htbp]
\centering
    \begin{tabular}{llll}
    \hline
    \textbf{strategy} &  del. & sub. & ins. \\
    \hline
    Random selection & \checkmark & \checkmark & \checkmark \\
    Phonetic similarity & & \checkmark & \\
    Linguistic probability & & & \checkmark \\
    \hline
    \end{tabular}
\caption{Hypothesis generation strategies and edit types. del.: deletion, sub.: substitution, ins.: insertion.}
\label{table:Hypothesis generation strategies and edit types}
\end{table}

\begin{table*}[htbp]
\centering
    \small
    \begin{tabular}{l|lllll lllll l}
    \hline
    \textbf{step} & \multicolumn{11}{c}{\textbf{text}} \\
    \hline
    reference     &on   &the  &     &morning&of   &september&eleventh&     &two  &thousand&and \\
    token(del/sub)&on   &the  &     &[sub]  &of   &[del]    &eleventh&     &two  &[sub]   &and \\
    replacement   &on   &the  &     &talking&of   &         &eleventh&     &two  &gunned  &and \\
    token(ins)    &on   &the  &[ins]&talking&of   &         &eleventh&[ins]&two  &gunned  &and \\
    hypothesis    &on   &the  &one  &talking&of   &         &eleventh&down &two  &gunned  &and \\
    \hline
    \end{tabular}
\caption{Example of hypothesis generation.}
\label{table:Example of hypothesis generation}
\end{table*}

\subsubsection{Hypothesis Generation Method}
\label{subsection:Hypothesis Generation Method}
\textbf{Random Sampling of Transcripts}
The first approach to hypothesis generation is to draw samples randomly from the transcripts of the ASR corpus. In other words, the spoken utterances and transcripts are paired randomly. This method guarantees that the vocabulary list, the number of words and the distribution of words in the training, validation and evaluation dataset do not change during hypothesis generation. 

\textbf{Random Sampling of Words}
Another random sampling can be at a word level. The words in a reference are replaced with the words randomly chosen according to the vocabulary distribution of the datasets. With this method, the total number of words and the distribution of words in the dataset can be maintained in addition to the length of a transcript. However, with the random sampling methods, the WER of the hypotheses can not be targeted. For example, hypotheses of 10\% of WER. To address these issues, a method for controlling WER is introduced in the following section. 

\textbf{Edit Generation}
In contrast to the random sampling methods, individual edits on transcripts can be generated to achieve a target WER for a dataset. WER is the ratio of the number of deletion, substitution and insertion errors in hypotheses to the number of words in references. As the reference does not change, the WER of a dataset can be controlled by the total number of errors.

To keep the number of edits as close to the target as possible, the tokens of deletion and substitution, \textit{[del]} and \textit{[sub]}, respectively, replace the words to be deleted or substituted. Then, the \textit{[del]} tokens are deleted from the reference and the \textit{[sub]} tokens are replaced by other words. For substitution, a phonetic similarity matrix is built using edit distance between phoneme sequences of two words. Then, \textit{[sub]} tokens could be substituted by phonetically similar ones, e.g., \textit{born} replaced by \textit{borne}. Among the words in the phonetic similarity matrix, a word for the substitution is randomly selected according to the similarity. 

The total number of each edit type could change whenever shorter paths are found as a result of the insertion. For example, a sequence of a deletion and an insertion converts into a substitution. To minimise the unexpected change in the number of individual edits, a token for insertion, \textit{[ins]}, is inserted between correct words after generating deletions and insertions. For the \textit{[ins]} token, it is replaced by linguistically probable ones, e.g., \textit{it} can be inserted into the end of \textit{because of}. In a similar way to the substitution, the word is drawn from the most probable word list. The probability is obtained by the language model trained on the reference transcripts of a training dataset. 

The example of the hypothesis generation is described in Table~\ref{table:Example of hypothesis generation}.

\subsection{Data augmentation}
Training data are augmented by merging hypothesis sets of different WERs. Each set is generated individually. When they are combined, some instances can be duplicated. To maintain the same amount of data generated, they are not removed. Therefore, if the same number of datasets are merged, the amount of training data will be the same.

\section{Experimental Setup}
\subsection{ASR Corpora}
Ted-Lium 3 (TL3) \cite{10.1007/978-3-319-99579-3_21} was used as a corpus for training WER estimators. This corpus has been used for the WER estimation task in previous studies \cite{10095888, park2023fast}. While the TL3 test dataset was used for in-domain evaluation, three evaluation datasets from different domains were used for the evaluation of the models on out-of-domain data. First, FCASC is an evaluation dataset for AMI \cite{kraaij2005ami}. AMI is a multiparty meeting corpus recorded in business meetings of three or four participants. These participants played the roles of employees in a business situation with or without scenarios. Second, 2000 HUB5 English Evaluation Speech \footnote{https://catalog.ldc.upenn.edu/LDC2002S09} is an English conversational telephone speech dataset. It consists of 20 unreleased conversations from the Switchboard study \cite{225858} as well as 20 conversations from CALLHOME American English Speech \footnote{https://catalog.ldc.upenn.edu/LDC97S42}. The Switchboard conversations are between two people on daily topics. The CALLHOME conversations are between family members or close friends. Finally, Wall Street Journal (WSJ) \cite{paul-baker-1992-design} is a read speech corpus based on mainly WSJ materials. One of the features of WSJ is the transcripts with verbalised punctuation, e.g., COMMA. For WSJ, there are multiple evaluation datasets with different vocabulary sizes: eval92 5k, eval92 20k, eval93 5k, eval93 20k. For the following experiment, the four evaluation datasets for WSJ were merged into one dataset. The evaluation datasets from different domains were named as AMI eval, SWB/CH eval and WSJ eval. 

\subsection{ASR Systems}
\label{subsection:ASR Systems}
Training data for WER estimation were generated by transcribing the TL3 train set through different ASR systems in Section~\ref{subsection:Automatic Speech Recognition Systems}: Whisper~\cite{radford2022robust}, wav2vec 2.0~\cite{NEURIPS2020_92d1e1eb}, Chain~\cite{povey2014parallel}, Conformer~\cite{gulati20_interspeech}, Transducer~\cite{graves2012sequence}. These models and their weights were downloaded from online resources\footnote{https://github.com/openai/whisper}\footnote{https://github.com/facebookresearch/fairseq}\footnote{https://github.com/kaldi-asr/kaldi}\footnote{https://huggingface.co/speechbrain}\footnote{https://github.com/speechbrain/speechbrain} except Chain. The Chain model was trained on 100 hours of LibriSpeech \cite{7178964} using LF-MMI~\cite{povey16_interspeech} and its augmented versions by changing speed and volume. The details of each model size and their training data are described in Table~\ref{table:Summary of automatic speech recognition systems}.
\begin{table}[htbp]
\small
    \begin{center}
    \begin{tabular}{llll}
    \hline
    & \textbf{model} & \textbf{trained on} & \begin{tabular}{@{}l@{}}\textbf{language}\\\textbf{model}\end{tabular} \\
    \hline
    ASR1 & Whisper     & 680k from net. & Transformer \\
    ASR2 & wav2vec 2.0 & LS 960h        & None        \\
    ASR3 & Chain       & LS 100h        & 3-gram      \\
    ASR4 & Conformer   & LS 960h        & Transformer \\
    ASR5 & Transducer  & LS 960h        & RNN         \\
    \hline
    \end{tabular}
    \end{center}
\caption{Summary of ASR systems.}
\label{table:Summary of automatic speech recognition systems}
\end{table}

\begin{table}[htbp]
        \begin{center}
        \begin{tabular}{lll}
        \hline
        & \textbf{methods and strategies} \\
        \hline                                      
        GEN1 & random sampling of transcript       \\
        GEN2 & random sampling of word\\
        GEN3 & edit generation           \\
        GEN4 & edit generation, PS10      \\
        GEN5 & edit generation, PS30      \\
        GEN6 & edit generation, PS50      \\
        GEN7 & edit generation, PS100      \\
        GEN8 & edit generation, PS100, LS100 \\
        \hline
        \end{tabular}
        \end{center}
\caption{Hypothesis generation methods. PS\textit{m}: phonetically similar \textit{m}  words, LS\textit{m}: linguistically similar \textit{m} words}
\label{table:Hypothesis generation methods}
\end{table}

\begin{figure}[htbp]
    \begin{minipage}[b]{0.38\linewidth}
    \centering
    \centerline{\includegraphics[height=4.0cm]{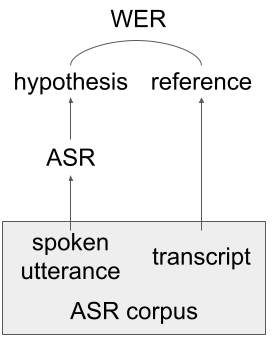}}
    \centerline{(a) ASR hypothesis}\medskip
    \end{minipage}
    \begin{minipage}[b]{0.68\linewidth}
    \centering
    \centerline{\includegraphics[height=4.0cm]{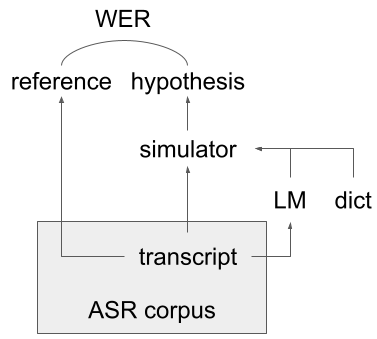}}
    \centerline{(b) Hypothesis generation}\medskip
    \end{minipage}
    \caption{Illustration of ASR hypothesis and Hypothesis generation}
    \label{figure:Illustration of ASR hypothesis and Hypothesis generation}
\end{figure}
\subsection{Training Datasets} 
\label{subsection:Training Datasets}
The training datasets for WER estimation were generated in two ways as described in Figure~\ref{figure:Illustration of ASR hypothesis and Hypothesis generation}: transcribing the TL3 train dataset with an ASR model and simulating the ASR output. First, the training dataset was transcribed using the ASR models in Section~\ref{subsection:ASR Systems}. To prevent imbalanced data from the low WER, the number of instances whose WER was 0 was limited to the sum of the second and the third common WER, e.g. the total number of instances in the range of WER of 3\% and 4\%. 

Second, the hypotheses were generated by hypothesis generation methods described in Section~\ref{subsection:Hypothesis Generation Method} and Table~\ref{table:Hypothesis generation methods}. For GEN1, a spoken utterance and a transcript were paired randomly and uniformly. For GEN2, each word was replaced with another word randomly sampled from the word distribution of the TL3 train dataset. GEN3 was a method to generate hypotheses aiming for target WERs of 2\%, 4\%...100\%. From GEN4 to GEN7, phonetic similarity was adopted, while the linguistic probability was employed by GEN8. When the target WER is given, the target number for each edit is distributed almost equally. For example, the WER of insertions, deletions and substitutions generated using GEN7 were 3.06\%, 3.33\% and 3.29\% when the target WER was 10\%. For substitution, the length of the phonetically similar word list was different from 10, 30, 50 and 100: GEN4 to GEN7, respectively. Moreover, for linguistic probability, a 3-gram language model was trained on the TL3 train with the SRI Language Modeling toolkit\footnote{http://www.speech.sri.com/projects/srilm/}. Additionally, several datasets were merged for data augmentation. For example, the generated hypotheses of 10\%, 20\%...100\% WER were merged into the dataset called GEN7W10-100 when the hypothesis generation method is GEN7. The datasets generated by ASR systems and hypothesis generations are organised in Table~\ref{table:Training datasets for WER estimation}. 
\begin{table}[htbp]
        \begin{center}
        \begin{tabular}{llll}
        \hline
        \textbf{dataset} & \textbf{hours} & \textbf{WER} & \begin{tabular}{@{}l@{}}\textbf{std.}\\\textbf{dev.}\end{tabular} \\
        \hline
        HYP1 train            & 256.22 & 0.1913 & 0.3184 \\
        HYP2 train            & 372.62 & 0.1618 & 0.1939 \\
        HYP3 train            & 434.78 & 0.3172 & 0.2419 \\
        HYP4 train            & 356.19 & 0.1580 & 0.2019 \\
        HYP5 train            & 376.52 & 0.1736 & 0.1864 \\
        \hline                                             
        GEN1 train            & 444.62 & 1.2898 & 2.9786 \\
        GEN2 train            & 444.62 & 0.9890 & 0.0320 \\
        GEN\textit{n}W\textit{m} train & 444.62 & approx. \textit{m} &  \\
        \hline
        \end{tabular}
        \end{center}
\caption{Training datasets for WER estimation. HYP\textit{n}: transcribed by ASR\textit{n}, GEN\textit{n}W\textit{m}: generated by GEN\textit{n} with target WER of \textit{m}}
\label{table:Training datasets for WER estimation}
\end{table}

\subsection{Evaluation Datasets}
The evaluation datasets were generated by the ASR systems in Table~\ref{table:Summary of automatic speech recognition systems}. The instances of 0 WER were filtered out as described in Section~\ref{subsection:Training Datasets}. As a result, the total duration of audio could be reduced differently according to an ASR system. The WER estimation models were also evaluated on the out-of-domain test sets, AMI eval, SWB eval and WSJ eval. Each test set was transcribed by ASR1\textendash5. WER of the evaluation datasets are summarised in Table~\ref{table:WER on evaluation datasets}. 

\begin{table}[htbp]
        \begin{center}
        \begin{tabular}{lllll}
        \hline
        \textbf{dataset} & \textbf{hrs.} & \textbf{WER} & \textbf{std.} \\
        \hline
        in-domain       &      &        &        \\
        HYP1 eval (TL3) & 1.97 & 0.0979 & 0.1935 \\
        HYP2 eval (TL3) & 3.41 & 0.1243 & 0.1501 \\
        HYP3 eval (TL3) & 4.18 & 0.2099 & 0.1806 \\
        HYP4 eval (TL3) & 3.38 & 0.1226 & 0.1673 \\
        HYP5 eval (TL3) & 3.53 & 0.1300 & 0.1445 \\
        \hline
        out-of-domain      &      &        &        \\
        HYP1 eval (AMI)    & 7.34 & 0.2754 & 0.4068 \\
        HYP2 eval (AMI)    & 8.16 & 0.3883 & 0.3542 \\
        HYP3 eval (AMI)    & 8.68 & 0.5946 & 0.3184 \\
        HYP4 eval (AMI)    & 8.18 & 0.3934 & 0.3555 \\
        HYP5 eval (AMI)    & 8.68 & 0.4087 & 0.3613 \\
        HYP1 eval (SWB/CH) & 2.75 & 0.1970 & 0.3522 \\
        HYP2 eval (SWB/CH) & 3.37 & 0.3336 & 0.3569 \\
        HYP3 eval (SWB/CH) & 3.56 & 0.6558 & 0.2889 \\
        HYP4 eval (SWB/CH) & 3.29 & 0.3412 & 0.3590 \\
        HYP5 eval (SWB/CH) & 3.56 & 0.6561 & 0.3575 \\
        HYP1 eval (WSJ)    & 2.13 & 0.0307 & 0.0748 \\
        HYP2 eval (WSJ)    & 1.60 & 0.1293 & 0.1166 \\
        HYP3 eval (WSJ)    & 2.22 & 0.1388 & 0.1332 \\
        HYP4 eval (WSJ)    & 1.50 & 0.1258 & 0.1123 \\
        HYP5 eval (WSJ)    & 1.62 & 0.1306 & 0.1120 \\
        \hline
        \end{tabular}
        \end{center}
\caption{WER on evaluation datasets.}
\label{table:WER on evaluation datasets}
\end{table}

\begin{table*}[htbp]
    \small
        \begin{center}
        \begin{tabular}{lllllllllll}
        \hline
        \multirow{4}{*}{\textbf{model}} & \multicolumn{10}{c}{\textbf{evaluated on}}\\
        \cmidrule(lr){2-11}
        & \multicolumn{2}{c}{\textbf{HYP1 eval}} & \multicolumn{2}{c}{\textbf{HYP2 eval}} & \multicolumn{2}{c}{\textbf{HYP3  eval}} & \multicolumn{2}{c}{\textbf{HYP4 eval}} & \multicolumn{2}{c}{\textbf{HYP5 eval}}\\
        \cmidrule(lr){2-3} \cmidrule(lr){4-5} \cmidrule(lr){6-7} \cmidrule(lr){8-9} \cmidrule(lr){10-11}
        & \textbf{RMSE$\downarrow$} & \textbf{PCC$\uparrow$} & \textbf{RMSE$\downarrow$} & \textbf{PCC$\uparrow$} & \textbf{RMSE$\downarrow$} & \textbf{PCC$\uparrow$} & \textbf{RMSE$\downarrow$} & \textbf{PCC$\uparrow$} & \textbf{RMSE$\downarrow$} & \textbf{PCC$\uparrow$} \\
        \hline
        Fe-WER1 & \textbf{0.0926} & \textbf{0.8806} & 0.1234 & 0.6524  & 0.1876 & 0.5375 & 0.1404 & 0.6658 & 0.1333 & 0.5491 \\
        Fe-WER2 & 0.1350 & 0.7210 & \textbf{0.0928} & \textbf{0.7962}  & 0.1494 & 0.6807 & 0.1151 & 0.7584 & 0.1069 & 0.6921 \\
        Fe-WER3 & 0.1404 & 0.7003 & 0.1139 & 0.7454  & \textbf{0.1148} & \textbf{0.7790} & 0.1166 & 0.7239 & 0.1165 & 0.6745 \\
        Fe-WER4 & 0.1275 & 0.7565 & 0.1084 & 0.7323  & 0.1459 & 0.6556 & \textbf{0.1090} & \textbf{0.7611} & 0.1125 & 0.6656 \\
        Fe-WER5 & 0.1402 & 0.6908 & 0.1017 & 0.7662  & 0.1303 & 0.7148 & 0.1105 & 0.7517 & \textbf{0.1064} & \textbf{0.6997} \\
        \hline
        \end{tabular}
        \end{center}
\caption{RMSE and PCC of WER estimators trained and evaluated on ASR hypotheses of TL3. Fe-WER\textit{n} is an estimator trained on HYP\textit{n} train.}
\label{table:RMSE and PCC of WER estimators trained and evaluated on ASR hypotheses of TL3}
\end{table*}

\begin{table*}[htbp]
    \small
        \begin{center}
        \begin{tabular}{lllllllllll}
        \hline
        \multirow{4}{*}{\textbf{model}} & \multicolumn{10}{c}{\textbf{evaluated on}}\\
        \cmidrule(lr){2-11}
        & \multicolumn{2}{c}{\textbf{HYP1 eval}} & \multicolumn{2}{c}{\textbf{HYP2 eval}} & \multicolumn{2}{c}{\textbf{HYP3  eval}} & \multicolumn{2}{c}{\textbf{HYP4 eval}} & \multicolumn{2}{c}{\textbf{HYP5 eval}}\\
        \cmidrule(lr){2-3} \cmidrule(lr){4-5} \cmidrule(lr){6-7} \cmidrule(lr){8-9} \cmidrule(lr){10-11}
        & \textbf{RMSE$\downarrow$} & \textbf{PCC$\uparrow$} & \textbf{RMSE$\downarrow$} & \textbf{PCC$\uparrow$} & \textbf{RMSE$\downarrow$} & \textbf{PCC$\uparrow$} & \textbf{RMSE$\downarrow$} & \textbf{PCC$\uparrow$} & \textbf{RMSE$\downarrow$} & \textbf{PCC$\uparrow$} \\
        \hline
        SIWE1 & 0.8730 & 0.0754 & 0.8447 & 0.1022 & 0.7776 & 0.0516 & 0.8494 & 0.0949 & 0.8416 & 0.0661 \\
        SIWE2 & 0.8797 & 0.1964 & 0.8507 & 0.2703 & 0.7845 & 0.2011 & 0.8436 & 0.0829 & 0.8484 & 0.1838 \\
        SIWE3 & 0.1542 & 0.6897 & 0.1320 & 0.5994 & 0.1558 & 0.6551 & 0.1626 & 0.5257 & 0.1331 & 0.5707 \\
        SIWE4 & 0.1471 & 0.7144 & 0.1222 & 0.6470 & 0.1488 & 0.6721 & 0.1583 & 0.5433 & 0.1315 & 0.5588 \\
        SIWE5 & 0.1472 & 0.7225 & 0.1235 & 0.6481 & 0.1503 & 0.6851 & 0.1621 & 0.5294 & 0.1324 & 0.5708 \\
        SIWE6 & 0.1484 & 0.7122 & 0.1252 & 0.6305 & 0.1495 & 0.6820 & 0.1612 & 0.5282 & 0.1307 & 0.5756 \\
        SIWE7 & 0.1443 & 0.7282 & 0.1195 & 0.6739 & 0.1479 & 0.6830 & 0.1598 & 0.5361 & 0.1283 & \textbf{0.5959} \\
        SIWE8 & \textbf{0.1268} & \textbf{0.7914} & \textbf{0.1133} & \textbf{0.7083} & \textbf{0.1369} & \textbf{0.7152} & \textbf{0.1455} & \textbf{0.6204} & \textbf{0.1266} & 0.5946\\
        \hline
        \end{tabular}
        \end{center}
\caption{RMSE and PCC of WER estimators trained on  GEN hypotheses and evaluated on ASR hypotheses of TL3. SIWE\textit{n} is trained on GEN\textit{n}W\textit{10\textendash100}.}
\label{table:RMSE and PCC of WER estimators trained on  GEN hypotheses and evaluated on ASR hypotheses of TL3}
\end{table*}

\subsection{WER Estimator}
MLPs are employed to predict WER. The layer sizes of the model were [2048, 600, 32, 1]. The outputs of hidden layers are dropped out at a rate of 0.1. The Adam optimiser is used with a learning rate of 0.007. The cosine scheduler with 15 iterations is used as an annealing factor of the learning rate. For feature extraction, HuBERT large~\cite{9585401} and XLM-R large~\cite{conneau-etal-2020-unsupervised} are adopted. The averaged representations are 1024-dimensional features. 

\section{Results}
\subsection{Evaluation on In-Domain Datasets}
\label{subsection:Evaluation on in-domain datasets}
The performance of the WER estimators trained on the ASR hypotheses is summarised in Table~\ref{table:RMSE and PCC of WER estimators trained and evaluated on ASR hypotheses of TL3}. The performance of the estimators was best when the evaluation dataset was generated with the same ASR used to transcribe the training dataset. For example, when the estimator was trained on HYP1 train (TL3), then the performance of the estimator was the best on HYP1 eval (TL3).

For ASR system-independent WER estimation, WER estimators were trained on the data augmented with the datasets in Table~\ref{table:Training datasets for WER estimation}. SIWE1 and SIWE2 are the estimation models trained on the GEN1 and GEN2 train, respectively. These random sampling methods for data generation showed relatively poor results. However, the RMSE and PCC of SIWE3\textendash8 are comparable to those of ASR system-dependent WER estimators, Fe-WER1\textendash5. SIWE8 outperformed the other SIWE models on most evaluation datasets.

For comparison between Fe-WERs and SIWEs, the RMSE and PCC values are averaged over HYP1\textendash5 eval without a duration weight, the mean RMSE and PCC of the SIWE8 model were better than Fe-WER1 by 0.0056 and 0.0289, respectively.
The results are shown in Table~\ref{table:RMSE and PCC of WER estimators trained on  GEN hypotheses and evaluated on ASR hypotheses of TL3}. 

\subsection{Evaluation on Out-of-Domain Datasets}
The estimators were evaluated on out-of-domain datasets in Table~\ref{table:WER on evaluation datasets}. For simplicity, the RMSE and PCC values are averaged over HYP1–5 eval without a duration weight. The mean RMSE and PCC are shown in Table~\ref{table:Mean RMSE and PCC of WER estimators evaluated on AMI, SWB/CH and WSJ eval}. In terms of the means of RMSE and PCC, the performance of SIWE7 and SIWE8 on AMI and SWB/CH eval are better than those of all Fe-WERs, while the results on WSJ eval are mixed. On WSJ eval, the RMSE and the PCC mean values of all the SIWE models are better than those of Fe-WER1, while they are worse than those of Fe-WER4. 

\begin{table*}[htbp]
\small
        \begin{center}
        \begin{tabular}{lllllllll}
        \hline
        \multirow{3}{*}{\textbf{model}} & \multicolumn{2}{c}{\textbf{AMI eval}} & \multicolumn{2}{c}{\textbf{SWB/CH eval}} & \multicolumn{2}{c}{\textbf{WSJ eval}} \\
        \cmidrule(lr){2-3} \cmidrule(lr){4-5} \cmidrule(lr){6-7}
                   & \textbf{RMSE$\downarrow$} & \textbf{PCC$\uparrow$} & \textbf{RMSE$\downarrow$} & \textbf{PCC$\uparrow$} & \textbf{RMSE$\downarrow$} & \textbf{PCC$\uparrow$} \\
        \hline
       Fe-WER1        & 0.3620      & 0.5705      & 0.3581      & 0.5356       & 0.2727      & 0.2623       \\
       Fe-WER2        & 0.4623      & 0.4523      & 0.4196      & 0.4653       & 0.1186      & \textbf{0.5274} \\
       Fe-WER3        & \textbf{0.3286}& \textbf{0.6097}& 0.3250      & \textbf{0.5645} & 0.2675      & 0.3944       \\
       Fe-WER4        & 0.5289      & 0.1741      & 0.4940      & 0.1601       & \textbf{0.1158}& 0.4926       \\
       Fe-WER5        & 0.3400      & 0.5978      & \textbf{0.3209}& 0.5601       & 0.1657      & 0.5060       \\
        \hline                                                                 
        SIWE7        & \textbf{0.2822} & \textbf{0.6764} & \textbf{0.2645} & \textbf{0.6673} & \textbf{0.2177} & 0.3628 \\
        SIWE8        & 0.2897 & 0.6518 & 0.2882 & 0.6224 & 0.2549 & \textbf{0.3947} \\
        \hline                                                                 
        \end{tabular}
        \end{center}
\caption{Mean RMSE and PCC of WER estimators evaluated on AMI, SWB/CH and WSJ eval.}
\label{table:Mean RMSE and PCC of WER estimators evaluated on AMI, SWB/CH and WSJ eval}
\end{table*}

\begin{table*}[htbp]
\small
        \begin{center}
        \begin{tabular}{lllllllll}
        \hline
        \multirow{3}{*}{\begin{tabular}{@{}l@{}}\textbf{WER}\\\textbf{range}\end{tabular}} & \multicolumn{2}{c}{\textbf{AMI eval}} & \multicolumn{2}{c}{\textbf{SWB/CH eval}} & \multicolumn{2}{c}{\textbf{WSJ eval}} \\
        \cmidrule(lr){2-3} \cmidrule(lr){4-5} \cmidrule(lr){6-7}
                   & \textbf{RMSE$\downarrow$} & \textbf{PCC$\uparrow$} & \textbf{RMSE$\downarrow$} & \textbf{PCC$\uparrow$} & \textbf{RMSE$\downarrow$} & \textbf{PCC$\uparrow$} \\
        \hline
        W2\textendash20   & 0.3454 & 0.6618 & 0.3243 & 0.6321 & \textbf{0.1092} & \textbf{0.4546} \\
        W12\textendash30  & 0.3106 & 0.6818 & 0.2886 & 0.6592 & 0.1193 & 0.4219 \\
        W22\textendash40  & 0.2957 & 0.6771 & 0.2714 & 0.6678 & 0.1452 & 0.3929 \\
        W32\textendash50  & 0.2835 & 0.6869 & 0.2636 & 0.6720 & 0.1673 & 0.3783 \\
        W42\textendash60  & 0.2765 & \textbf{0.6900} & \textbf{0.2593} & \textbf{0.6731} & 0.1961 & 0.3871 \\
        W52\textendash70  & \textbf{0.2747} & 0.6827 & 0.2642 & 0.6657 & 0.2610 & 0.3682 \\
        W62\textendash80  & 0.2755 & 0.6844 & 0.2673 & 0.6663 & 0.3063 & 0.3700 \\
        W72\textendash90  & 0.2803 & 0.6811 & 0.2746 & 0.6651 & 0.3191 & 0.4036 \\
        W82\textendash100 & 0.3023 & 0.6870 & 0.3125 & 0.6627 & 0.4588 & 0.3704 \\
        \hline                                                                 
        \end{tabular}
        \end{center}
\caption{Mean RMSE and PCC of SIWE7 trained on the datasets comprising different target WER ranges. Mean WERs of HYP1\textendash5 eval of AMI, SWB/CH and WSJ are 0.4121, 0.4367 and 0.1110, respectively.}
\label{table:Mean RMSE and PCC of WER estimators trained on the datasets comprising different target WER ranges}
\end{table*}

\section{Discussion}
\subsection{Data Generation Method}
In this section, we will discuss the effect of each generation method for WER estimation on in-domain and out-of-domain data in detail. First, the edit generation method, GEN3, improved the performance of the estimators significantly compared to GEN1 and GEN2. In Table~\ref{table:RMSE and PCC of WER estimators trained on  GEN hypotheses and evaluated on ASR hypotheses of TL3}, the mean RMSE of SIWE3 was 0.1542, while those of SIWE1 and SIWE2 were 0.8730 and 0.8797, respectively. The amount of the training data would be helpful to improve the performance of the estimator because the training dataset for SIWE3 was the merged dataset of GEN3W\textit{10\textendash100}. Moreover, all the PCC values of the estimators trained on GEN3\textendash7 were over 0.5, while those of SIWE1 and SIWE2 were below 0.3. Second, using phonetic similarity for substitution also brought an additional gain in the performance of the estimators. Additionally, as the size of the phonetically similar word list increased from 10 to 100, RSME and PCC of SIWE improved. Furthermore, when the linguistic probability was added to the generation method, GEN8, the performance was notably improved on all evaluation datasets. For example, the mean RMSE and PCC of the SIWE8 estimator on HYP1\textendash5 eval were 0.1298 and 0.6860, while those of the SIWE7 estimator were 0.1400 and 0.6434, respectively. 

However, the performance improvement by GEN8 could not be achieved on out-of-domain data. The results in Table~\ref{table:Mean RMSE and PCC of WER estimators evaluated on AMI, SWB/CH and WSJ eval} show that all the results of SIWE8 are behind the SIWE3\textendash7 on AMI and SWB/CH evals. These results mean that the linguistic information obtained from the TL3 corpus might not be helpful for WER estimation on different corpora. 

\subsection{Data Augmentation}
The performance of SIWEs was behind that of Fe-WERs on out-of-domain data except for WSJ eval. For further analysis, SIWEs were trained on the datasets consisting of different ranges of target WER. For example, training data of WER of 2\%, 4\%...20\%. The amount of data does not change from the datasets used in Table~\ref{table:Mean RMSE and PCC of WER estimators evaluated on AMI, SWB/CH and WSJ eval} as the same number of datasets were merged. The performance improvement was observed when the WER range of the training dataset was close to the WER of evaluation datasets. For example, SIWE7W2\textendash20 performed better than the others in terms of both mean RMSE and PCC when the mean WER of HYP1\textendash5 eval (WSJ) was 0.1110. Similarly, SIWE7W42\textendash60 outperformed the others on SWB/CH eval whose mean of WER on HYP1\textendash5 eval (SWB/CH) was 0.4367. Moreover, those results in Table~\ref{table:Mean RMSE and PCC of WER estimators trained on the datasets comprising different target WER ranges} are better than those of SIWE7 in Table~\ref{table:Mean RMSE and PCC of WER estimators evaluated on AMI, SWB/CH and WSJ eval}. 

\section{Conclusion}
Hypothesis generation and data augmentation for ASR system-independent WER estimation are proposed in this paper. The estimator trained on the hypotheses generated by this proposed method outperforms the baselines on out-of-domain datasets. As a hypothesis generation strategy, linguistic information helps improve the performance of WER estimators when they are evaluated on in-domain datasets. The phonetic similarity for substitution improved the performance of the estimators on both in-domain and out-of-domain datasets. On out-of-domain datasets, the performance of estimators has further improved marginally when the hypotheses are generated with a target WER close to the evaluation dataset. 

\section{Acknowledgements}
\label{sec:Acknowledgements}
This work was conducted at the Voicebase/Liveperson Centre of Speech and Language Technology at the University of Sheffield which is supported by Liveperson, Inc..

\section{Bibliographical References}\label{sec:reference}
\small
\bibliographystyle{lrec-coling2024-natbib}
\setlength{\bibsep}{2pt} 
\bibliography{lrec-coling2024}

\end{document}